\documentclass[letterpaper, 10 pt, conference]{ieeeconf}  

\IEEEoverridecommandlockouts                              

\usepackage[colorlinks,linkcolor=red,anchorcolor=blue,citecolor=green]{hyperref}
\usepackage{graphics} 
\usepackage{epsfig} 
\usepackage{times} 
\usepackage{cite}
\usepackage{amsmath} 
\usepackage{amssymb}  
\usepackage{comment}
\usepackage{multirow}
\usepackage{booktabs}
\usepackage{siunitx}

\title{\LARGE \bf
ACORN: Adaptive Contrastive Optimization for Safe and Robust Fine-Grained Robotic Manipulation
}

\author{Zhongquan Zhou$^{1}$ Shuhao Li$^{1}$ Zixian Yue $^{1}$
\thanks{$^{1}$School of Information Science \& Engineering, Lanzhou University, China.}%
}

\begin{document}

\maketitle
\thispagestyle{empty}
\pagestyle{empty}

\begin{abstract}

Embodied AI research has traditionally emphasized performance metrics such as success rate and cumulative reward, overlooking critical robustness and safety considerations that emerge during real-world deployment. In actual environments, agents continuously encounter unpredicted situations and distribution shifts, causing seemingly reliable policies to experience catastrophic failures, particularly in manipulation tasks. To address this gap, we introduce four novel safety-centric metrics that quantify an agent's resilience to environmental perturbations. Building on these metrics, we present \textbf{A}daptive \textbf{C}ontrastive \textbf{O}ptimization for \textbf{R}obust Ma\textbf{n}ipulation (ACORN), a plug-and-play algorithm that enhances policy robustness without sacrificing performance. ACORN leverages contrastive learning to simultaneously align trajectories with expert demonstrations while diverging from potentially unsafe behaviors. Our approach efficiently generates informative negative samples through structured Gaussian noise injection, employing a double perturbation technique that maintains sample diversity while minimizing computational overhead. Comprehensive experiments across diverse manipulation environments validate ACORN's effectiveness, yielding improvements of up to 23\% in safety metrics under disturbance compared to baseline methods. These findings underscore ACORN's significant potential for enabling reliable deployment of embodied agents in safety-critical real-world applications.

\end{abstract}

\section{Introduction}

Modern robotic manipulation systems face critical challenges in maintaining operational safety and robustness, which are essential requirements for real-world deployment in unstructured environments. While Action Chunking with Transformers (ACT) \cite{aloha} algorithm demonstrates impressive performance in fine-grained manipulation tasks, these methods exhibit fundamental vulnerabilities to environmental perturbations and distribution shifts that significantly limit their practical applicability. Traditional imitation learning approaches predominantly optimize for task completion metrics (e.g., success rates, reward maximization), often overlooking critical safety considerations that prove essential in dynamic real-world scenarios. This oversight becomes particularly consequential in safety-critical applications , where millimeter-scale trajectory deviations can trigger catastrophic failures i.

Current approaches to robustness enhancement face three primary limitations: 1) inadequate safety quantification in precision tasks, 2) computational inefficiency in perturbation handling, and 3) costly training sample collection and processing. Existing methods like Generative Adversarial Imitation Learning (GAIL) \cite{GAIR} and Reinforcement Learning with Augmented Data (RAD) \cite{RLD} attempt to address distribution shifts through data augmentation but remain constrained by inherent model dependencies. Contrastive learning approaches such as SimCLR \cite{SimCLR} show potential for trajectory alignment, yet their computational complexity and indirect safety optimization limit practical deployment in real-time systems. 

To address these limitations, we introduce ACORN, an adaptive contrastive optimization designed for robust and safe robotic manipulation. The key contributions of our work are as follows:

\begin{enumerate}
    \item \textbf{Safety-Centric Metric Ecosystem:} We introduce novel metrics to quantify safety in fine-grained manipulation tasks, including failure-conditional rewards and kinematic divergence measures, which provide deeper insights into the agent's resilience to perturbations.
    \item \textbf{Dual-Perturbation Contrastive Learning:} We propose a new contrastive learning strategy that generates negative samples through correlated scaling and independent micro-perturbations, enabling the agent to learn from both expert demonstrations and noisy data without computational overhead.
    \item \textbf{Biomechanical Priority Awareness:} We optimize safety by focusing on high-priority joints, which are critical for precision manipulation. 
    \item \textbf{Curriculum-Aware Robustness:} ACORN incorporates a progressive training strategy that adjusts contrastive loss weights based on policy maturity, preventing premature convergence and ensuring stable learning.
\end{enumerate}

\section{Related Work}

\subsection{Safety in Robot Learning}

Modern society increasingly relies on robotic systems. Therefore, designing safe learning methods for real-world robotic deployments is a critical research topic. Nevertheless, in many manipulation tasks, such as employing mobile manipulators for service applications \cite{dong2020catch} , the dynamics of these complex systems are often uncertain. For instance, sensor measurements can be noisy. The existing literature on the safety of robot learning can be categorized into two main approaches. The first approach focuses on learning uncertain dynamics to certify safety. These methods typically depend on prior knowledge of system dynamics \cite{liu2022robot, dawson2023safe} or on learning the uncertainties in the dynamics \cite{choi2020reinforcement, chen2021context, as2022constrained}. The second approach promotes safety during policy exploration or penalizes dangerous actions in a model‐free manner \cite{ke2021grasping}.Related work includes \cite{ke2023ccil, deshpande2024data}, which enhances the security of robot learning by incorporating data augmentation techniques that address noise. Continuity-based Data Augmentation for Corrective Imitation Learning \cite{ke2023ccil} relies on the assumption of local continuity, but our method can be seamlessly integrated into any variant of ACT algorithm without precondition and the underlying algorithmic approach is applicable to any imitation learning method. In \cite{deshpande2024data}, the focus is solely on improving success rate, with no consideration for safety.

\subsection{Imitation Learning}
Imitation learning (IL) \cite{osa2018algorithmic} is a widely used approach for learning an optimal policy when the reward function is unavailable \cite{ng2000algorithms}. In IL, an expert provides demonstrations of the desired behavior to train a policy. IL-derived policies enable computationally efficient online execution \cite{hertneck2018learning,yin2021imitation}. A substantial body of work has demonstrated the theoretical and practical advantages of IL‐based methods across various applications, including video game playing \cite{ross2011reduction,ross2010efficient}, humanoid robotics \cite{schaal1999imitation}, and autonomous driving \cite{codevilla2018end}. In this work, we extend the ACT algorithm to handle disturbed environment, thereby enhancing its applicability beyond the original formulation. Moreover, our plug-and-play approach can be integrated into various imitation learning frameworks to improve their robustness.

\subsection{Distribution Shift}

In embodied AI, distribution shift arises from discrepancies between training and real-world deployment environments, often due to mismatched physical dynamics\cite{grune2015models}, sensory noise, human biases during robot training \cite{thomaz2006reinforcement,thomaz2008teachable}, and action execution variances, significantly hindering agents' generalization and operational robustness. Despite its prevalence in real-world applications, most interactive imitation learning studies assume a stationary expert policy \cite{likmeta2021dealing,Shin_Lee_Yoo_Woo_Kim_Woo,Zheng_Verma_Zhou_Tsang_Chen_2021}. Our adaptive contrastive optimization imitation learning approach bridges this gap by introducing an optimization strategy that integrates contrastive and curriculum learning to accommodate expert distribution shift.

\section{Preliminaries}

\subsection{ACT Algorithm}

The ACT algorithm addresses the compounding error problem in imitation learning for fine-grained manipulation by predicting action sequences instead of single-step actions. Given an observation \( \mathbf{o}_t \) at time \( t \), ACT models a policy \( \pi_\theta(\mathbf{a}_{t:t+k} | \mathbf{o}_t) \) that outputs a sequence of \( k \) future actions \( \mathbf{a}_{t:t+k} \), effectively reducing the task horizon by a factor of \( k \). To handle multi-modal human demonstrations, ACT employs a Conditional Variational Autoencoder (CVAE) framework, where a latent variable \( \mathbf{z} \sim \mathcal{N}(0, \mathbf{I}) \) captures demonstration variability. The encoder \( q_\phi(\mathbf{z} | \mathbf{a}_{t:t+k}, \mathbf{o}_t) \) infers \( \mathbf{z} \) from demonstration data, while the decoder (policy) \( \pi_\theta(\mathbf{a}_{t:t+k} | \mathbf{o}_t, \mathbf{z}) \) generates action sequences conditioned on \( \mathbf{z} \). The training objective minimizes:
\begin{equation}
\begin{aligned}
    & \mathcal{L}_{\text{ACT}} = \mathcal{L}_{\text{reconst}} + \mathcal{L}_{\text{reg}} \\
    & \mathcal{L}_{\text{reconst}} = \mathbb{E}_{\mathbf{z} \sim q_\phi} \left[ \left\| \mathbf{a}_{t:t+k} - \pi_\theta(\mathbf{o}_t, \mathbf{z}) \right\|_1 \right] \\
    & \mathcal{L}_{\text{reg}} = \lambda \cdot D_{\text{KL}}\left( q_\phi(\mathbf{z} | \mathbf{a}_{t:t+k}, \mathbf{o}_t) \, \| \, \mathcal{N}(0, \mathbf{I}) \right)
\end{aligned}
\end{equation}

where \( \lambda \) balances reconstruction and KL regularization. At inference, \( \mathbf{z} \) is set to the prior mean (zero).

\subsection{Contrastive Learning}

Contrastive learning aims to learn representations by maximizing similarity between positive pairs while minimizing similarity between negative pairs in an embedding space. In the context of embodied AI, contrastive learning guides agents to align their execution trajectories with expert demonstrations while diverging from suboptimal or noisy trajectories. The formulation of a max margin contrastive loss can be articulated as follows:  
\begin{equation}
   \mathcal{L}_{\text{contrastive}} = \max\left( 0, \underbrace{\mathcal{D}(\tau_i, \tau_i^+)}_\text{positive distance} - \underbrace{\mathcal{D}(\tau_i, \tau_i^-)}_\text{negative distance} + \alpha \right) 
\end{equation}
where \( \tau_i \) denotes the agent's trajectory, \( \tau_i^+ \) and \( \tau_i^- \) represent positive (expert) and negative (noisy) trajectories, respectively, \( \mathcal{D}(\cdot,\cdot) \) measures trajectory dissimilarity (e.g., L2 distance), and \( \alpha > 0 \) enforces a margin to separate positive and negative pairs. By minimizing \( \mathcal{L}_{\text{contrastive}} \), the agent's policy generates trajectories that are discriminative against perturbations and generalize robustly to unseen environments.

\section{Methodology}

\begin{figure*}[!ht]
    \centering
    \includegraphics[width=\textwidth]{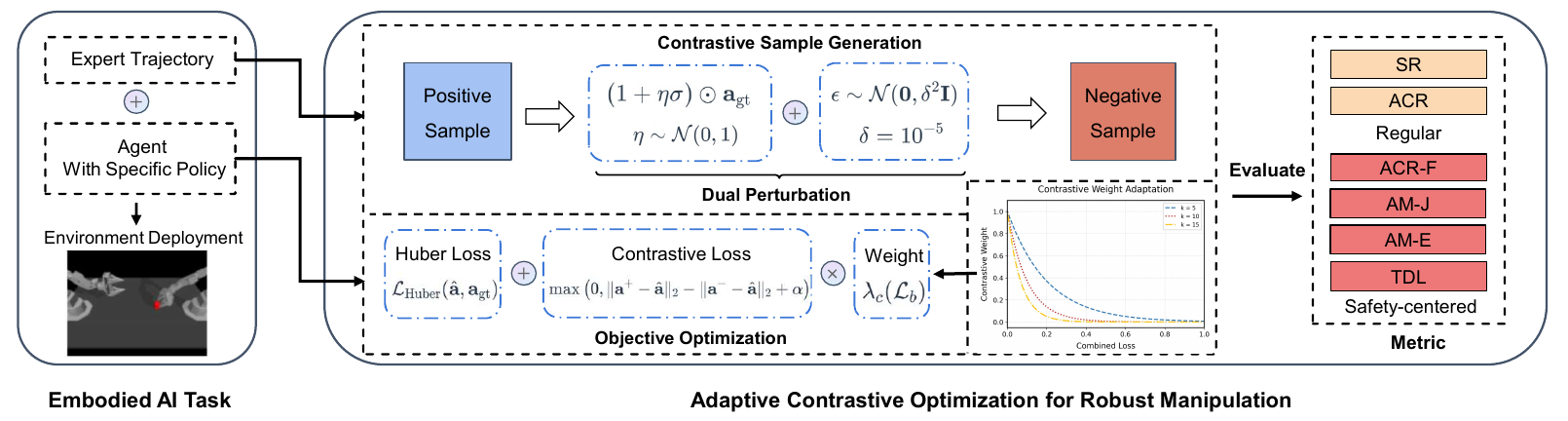}
    \caption{ACORN Overview: A three-stage optimization strategy for safety-aware reinforcement learning.}
    \label{fig:overview}
\end{figure*}

ACORN is purposed as a plug-and-play optimization to exhibit catastrophic failures under distributional shifts, especially in high-precision embodied operational tasks, which introduces a three-stage optimization paradigm that systematically enhances safety and robustness through progressive policy refinement. As shown in Fig~\ref{fig:overview}, ACORN systematically addresses safety-critical challenges by 1) maintaining baseline task performance while 2) improving resilience against distribution shifts and novel operational scenarios via a structured three-stage architecture.

\textbf{Stage I: Contrastive Sample Generation}

Our dual-perturbation mechanism synthesizes informative negative samples through: 1) correlated parameter scaling that maintains temporal consistency, and 2) independent micro-perturbations introducing controlled stochasticity to critical joint parameters.  

\textbf{Stage II: Objective Optimization}

Contrastive samples from Stage I are fed into a hybrid optimization module that augments the baseline loss function with with 1) Huber Loss and 2) Contrastive Margin Loss. Meanwhile, a novel curriculum learning scheduler dynamically adjusts the contrastive loss weight based on policy maturation.

\textbf{Stage III: Safety-Centric Evaluation}

Four safety-centric evaluation metrics provide quantitative measures of agent safety performance beyond traditional success rates. These metrics form a quantitative assessment framework, detailed in the \ref{metrics} section.

\subsection{Safety-Centric Evaluation Metrics}\label{metrics}

Traditional embodied AI benchmarks primarily employ Success Rate (SR) and Average Cumulative Reward (ACR) as evaluation criteria. While these metrics provide basic performance assessment, they fail to quantify safety-critical behaviors when addressing core challenges: 1) action sequencing under distribution shift and 2) error accumulation in long-horizon policies. To bridge this gap, we propose four safety-centric metrics:

\begin{enumerate}
    \item \textbf{Average Cumulative Reward for Failure (ACR-F)}: Quantifies the agent's ability to maintain near-optimal actions before catastrophic failures.  
    \begin{equation}
         \text{ACR-F} = \frac{1}{|\mathcal{F}|} \sum_{\tau_i \in \mathcal{F}} \sum_{t=0}^{T_i} r_t^{(i)} 
    \end{equation}
   where $\mathcal{F} = \{\tau_i | \mathbb{I}_{\text{success}}(\tau_i) = 0\}$. Higher ACR-F indicates better adherence to safe operation principles prior to failure.

   \item \textbf{Average Module of Joint Pose Vector (AM-J):} Measures deviation from safe joint trajectories in training data \( \mathcal{D}_{\text{train}} \). 
   \begin{equation}
       \text{AM-J} = \frac{1}{|\mathcal{F}|} \sum_{\tau_i \in \mathcal{F}} \frac{1}{T_i} \sum_{t=0}^{T_i} \| \mathbf{a}_t^{(i)} - \mathbf{a}_{\text{ref},t}^{(i)} \|_2
   \end{equation}
   where \( \mathbf{a}_{\text{ref},t}^{(i)} = \arg\min_{\mathbf{a}'_t \in \mathcal{D}_{\text{train}}} \| \mathbf{a}_t^{(i)} - \mathbf{a}'_t \|_2 \). Lower AM-J implies better compliance with learned safe joint movements.

   \item \textbf{Average Module of End-Effector Vector (AM-E):} Evaluates end-effector trajectory safety through positional divergence. 
   \begin{equation}
       \text{AM-E} = \frac{1}{|\mathcal{F}|} \sum_{\tau_i \in \mathcal{F}} \frac{1}{T_i} \sum_{t=0}^{T_i} \| \mathbf{p}_t^{(i)} - \mathbf{p}_{\text{ref},t}^{(i)} \|_2
   \end{equation}
   with \( \mathbf{p}_{\text{ref},t}^{(i)} \) derived similarly to \( \mathbf{a}_{\text{ref},t}^{(i)} \). This metric is particularly crucial for gripper-based manipulation tasks.

   \item \textbf{Trajectory Departure Level (TDL)}: Visualizing violations of security boundaries through trajectory discrete analysis. We construct confidence regions based on expert trajectories in \(\mathcal{D}_{\text{train}}\) and then visualize failed event states.
   
\end{enumerate}

Our experimental results in Section \ref{experimants} demonstrate their effectiveness in diagnosing safety-critical failures that traditional metrics overlook.

\subsection{Noise Resilience Evaluation}  
A Gaussian noise injection scheme with probabilistic activation and scaled intensity is implemented to evaluate robustness. At each control step, noise triggering is probabilistically governed by parameter \( p \). When activated, the perturbation magnitude is computed as \(\beta = \eta \cdot \sigma\), where \(\eta \sim \mathcal{N}(0,1)\) and \(\sigma\) denotes the predefined intensity. The perturbed action is then defined as: 
\begin{equation}  
    \mathbf{a}_{\text{noised}} =   
   \begin{cases}   
   (1 + \beta) \mathbf{a}_{\text{gt}}, & \text{if activated (prob. } p) \\  
   \mathbf{a}_{\text{gt}}, & \text{otherwise}  
   \end{cases}  
\end{equation}  
where \(\mathbf{a}_{\text{gt}}\) represents the ground-truth action. Three noise levels are systematically evaluated: \textbf{LIGHT} (\(p = 0.1, \sigma = 0.04\)), \textbf{NORMAL} (\(p = 0.2, \sigma = 0.06\)), and \textbf{HEAVY} (\(p = 0.3, \sigma = 0.08\)).

\subsection{Contrastive Sample Generation via Dual Perturbation}

In conventional contrastive learning for embodied AI, manual sample curation is limited by labor-intensive annotation requirements. To address this, an automated dual-perturbation mechanism is proposed in ACORN, which achieves three key advantages: 1) elimination of human annotation costs, 2) preservation of policy learning efficacy, and 3) enforcement of safety-aware trajectory constraints.  

Negative samples are synthesized through correlated scaling and independent infinitesimal distortions:
\begin{equation}
    \mathbf{a}^- = \underbrace{(1 + \eta\sigma) \odot \mathbf{a}_{\text{gt}}}_{\text{Correlated Scaling}} + \underbrace{\epsilon}_{\text{Independent Perturbation}}
\end{equation}

where the perturbation parameters are configured as:
\begin{align*}
    \eta &\sim \mathcal{N}(0, 1), \quad \sigma  \in [0.04,0.06,0.08]\\
    \epsilon &\sim \mathcal{N}(\mathbf{0}, \delta^2\mathbf{I}), \quad \delta = 10^{-5}
\end{align*}

\subsection{ACT-ACORN Objective Function}
\label{ssec:act_acorn_objective}

The ACT-ACORN extends the original ACT objective through two fundamental enhancements:

\begin{itemize}
    \item \textbf{Robust Regression:} L1 loss is replaced by an adaptive Huber loss with optimized thresholding.
    \item \textbf{Curriculum-Guided Contrastive Learning}: Trajectory discrimination loss is dynamically modulated via progress-aware weighting.
\end{itemize}

The composite objective function integrates these components as follows:
\begin{equation}
    \mathcal{L}_{\text{ACT-ACORN}} = \mathcal{L}_{\text{Huber}} + \lambda_{\text{KL}} \mathcal{L}_{\text{KLD}} + \lambda_c(\mathcal{L}_b) \mathcal{L}_{\text{Contrast}}
    \label{eq:composite_loss}
\end{equation}

where $\mathcal{L}_b = \mathcal{L}_{\text{Huber}} + \lambda_{\text{KL}}\mathcal{L}_{\text{KLD}}$ denotes the policy’s baseline loss.

\subsubsection{Huber Loss} 
The Huber loss implements dual regimes for robust regression:
\begin{equation}
   \mathcal{L}_{\text{Huber}}(\hat{\mathbf{a}},\mathbf{a}_{\text{gt}}) = \begin{cases} 
       \frac{1}{2}(\mathbf{\hat{a}} - \mathbf{a}_{\text{gt}})^2, & |\mathbf{\hat{a}} - \mathbf{a}_{\text{gt}}| \leq \delta \\
       \delta |\mathbf{\hat{a}} - \mathbf{a}_{\text{gt}}| - \frac{1}{2}\delta^2, & \text{otherwise}
   \end{cases}
   \label{eq:huber}
\end{equation}

The threshold $\delta$ is a hyper-parameter set to $0.124$ based on empirical analysis of pretraining loss dynamics, representing the mean loss plus one standard deviation. This value optimally balances gradient stability for small residuals while maintaining robustness against outliers.
\subsubsection{Contrastive Loss} 

The contrastive loss is implemented with a hinge-style penalty on positive-negative pairs to enforce a margin-separated embedding space, optimizing similarity discrimination.
\begin{equation}
    \mathcal{L}_{\text{Contrast}} = \max\left(0, \|\mathbf{a}^+ - \hat{\mathbf{a}}\|_2 - \|\mathbf{a}^- - \hat{\mathbf{a}}\|_2 + \alpha \right)
    \label{eq:contrastive}
\end{equation}

where $\mathbf{a}^+$ and $\mathbf{a}^-$ denote positive and negative action samples, respectively. The margin parameter $\alpha = 0.01$ was optimized through systematic ablation studies to maximize discriminative power while maintaining training stability.

\subsubsection{Curriculum-Based Weight Adaptation} 

\begin{figure}[!ht]
    \centering
    \includegraphics[width=0.45\textwidth]{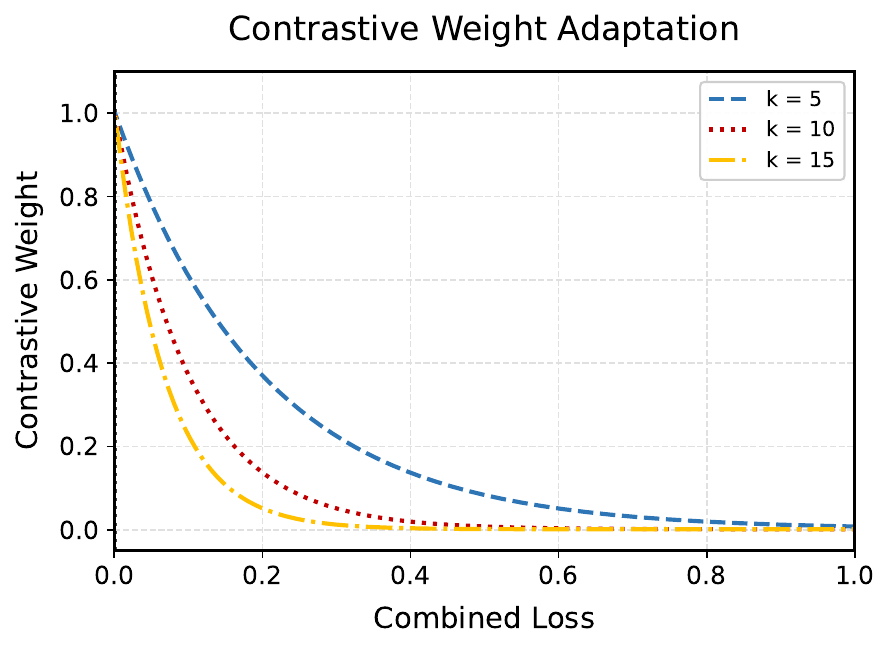}
    \caption{The contrastive loss contribution increases exponentially as baseline loss $\mathcal{L}_b$ decreases below unity.}
    \label{fig:contrastive_weight}
\end{figure}

Curriculum Learning (CL) is a training strategy inspired by human education, where models are progressively trained from easier to harder data samples to enhance generalization and convergence \cite{bengio2009curriculum}. As visualized in Fig ~\ref{fig:contrastive_weight}, the contrastive loss weight $\lambda_c$ evolves through three training phases governed by baseline loss progression:

\begin{equation*}
    \lambda_c(\mathcal{L}_b) = \begin{cases}
        10^{-3}, & \mathcal{L}_b > 1 \\
        1 - 0.999(1 - e^{-k(\mathcal{L}_b - 10^{-3})}), & 10^{-3} \leq \mathcal{L}_b \leq 1 \\
        1, & \mathcal{L}_b < 10^{-3}
    \end{cases}
\end{equation*}

The exponential coefficient $k$ is a hyper-parameter set to $15$ based on ablation experiments, this value maximizes the effect of the curriculum learning mechanism.

\section{Experiments}\label{experimants}

\subsection{Framework and Implementation}

Our implementation leverages the LeRobot framework (v2.0) \cite{cadene2024lerobot}. The baseline model adopts the ACT policy pretrained on the AlohaTransferCube task, with the following key specifications:

\begin{itemize}
\item \textbf{Sensors}: Top-mounted RGB camera (480×640×3 @50Hz) + 14D proprioceptive vectors.
\item \textbf{Actuators}: 14D impedance-controlled motor commands (6DOF×2 arms)
\item \textbf{Dataset}: 50 human demonstrations of cube manipulation
\end{itemize}

The ACORN extension maintains architectural compatibility with ACT while introducing contrastive learning objectives. Key implementation details include:

\begin{itemize}
\item Batch size increased from 8 to 32 for stable contrastive learning
\item 7.5hr training on NVIDIA RTX 3090 with original hyperparameters
\item Evaluation protocol: 500 episodes per noise condition (Light/Normal/Heavy)
\end{itemize}

The implementation preserves compatibility with upstream LeRobot features including real-time policy deployment and multi-camera sensor fusion, establishing a foundation for future sim-to-real transfer experiments.

\subsection{Main Results}

We conducted comprehensive evaluations comparing ACT-ACORN with the baseline ACT method under three noise levels. Table~\ref{tab:comparison_results} reveals three key advantages of our approach:

\begin{table}[!ht]
\centering
\caption{Comparative Performance Analysis: ACT vs. ACT-ACORN}
\label{tab:comparison_results}
\small
\begin{tabular}{l l c c}
\toprule
\textbf{Metric} & \textbf{Noise} & \textbf{ACT} & \textbf{ACT-ACORN} \\
\midrule
\multirow{3}{*}{SR (\%)} 
    & Light       & \textbf{80.80}   & 77.40            \\
    & Normal      & \textbf{62.20}   & \textbf{62.20}   \\
    & Heavy       & 22.80            & \textbf{23.00}   \\
\cmidrule(r){1-4}

\multirow{3}{*}{ACR} 
    & Light       & \textbf{220.252} & 211.704          \\
    & Normal      & 176.182          & \textbf{179.872} \\
    & Heavy       & 94.520           & \textbf{99.278}  \\
\cmidrule(r){1-4}

\multirow{3}{*}{ACR-F} 
    & Light       & \textbf{70.031}  & 66.301           \\
    & Normal      & 69.952           & \textbf{86.116}  \\
    & Heavy       & 60.085           & \textbf{67.182}  \\
\cmidrule(r){1-4}

\multirow{3}{*}{AM-J} 
    & Light       & 0.288            & \textbf{0.261}   \\
    & Normal      & 0.284            & \textbf{0.273}   \\
    & Heavy       & \textbf{0.253}   & 0.293            \\
\cmidrule(r){1-4}

\multirow{3}{*}{AM-E} 
    & Light       & 0.145            & \textbf{0.128}   \\
    & Normal      & 0.136            & \textbf{0.123}   \\
    & Heavy       & 0.116            & \textbf{0.115}   \\
\bottomrule
\end{tabular}
\end{table}

\begin{itemize}
\item \textbf{Superior Fine Control:} Under Normal noise, ACT-ACORN achieves \textbf{23.1\%} higher ACR-F (86.12 vs 69.95), demonstrating enhanced precision in delicate manipulation tasks
\item \textbf{Energy Efficiency:} Reduces motion energy consumption by \textbf{9.6\%} (AM-E: 0.123 vs 0.136) while maintaining equivalent success rates
\item \textbf{Noise Robustness:} Shows progressive improvements in heavy noise conditions (\textbf{+5.0\%} ACR, \textbf{+11.8\%} ACR-F)
\end{itemize}

Building upon established principles of human upper limb motor control, we propose a biomechanically-informed taxonomy that partitions the robotic arm's 6-DOF configuration into two functionally specialized groups:

\begin{itemize}
\item \textbf{High-Priority Joint Group}: Shoulder, Elbow, and Forearm Roll joints, identified as critical for precision manipulation based on their kinematic influence on end-effector positioning.
\item \textbf{Secondary Joint Group}: Waist, Wrist Angle, and Wrist Rotation joints, responsible for gross positioning and orientation adjustments
\end{itemize}

\begin{figure}[h]
\centering
\includegraphics[width=\linewidth]{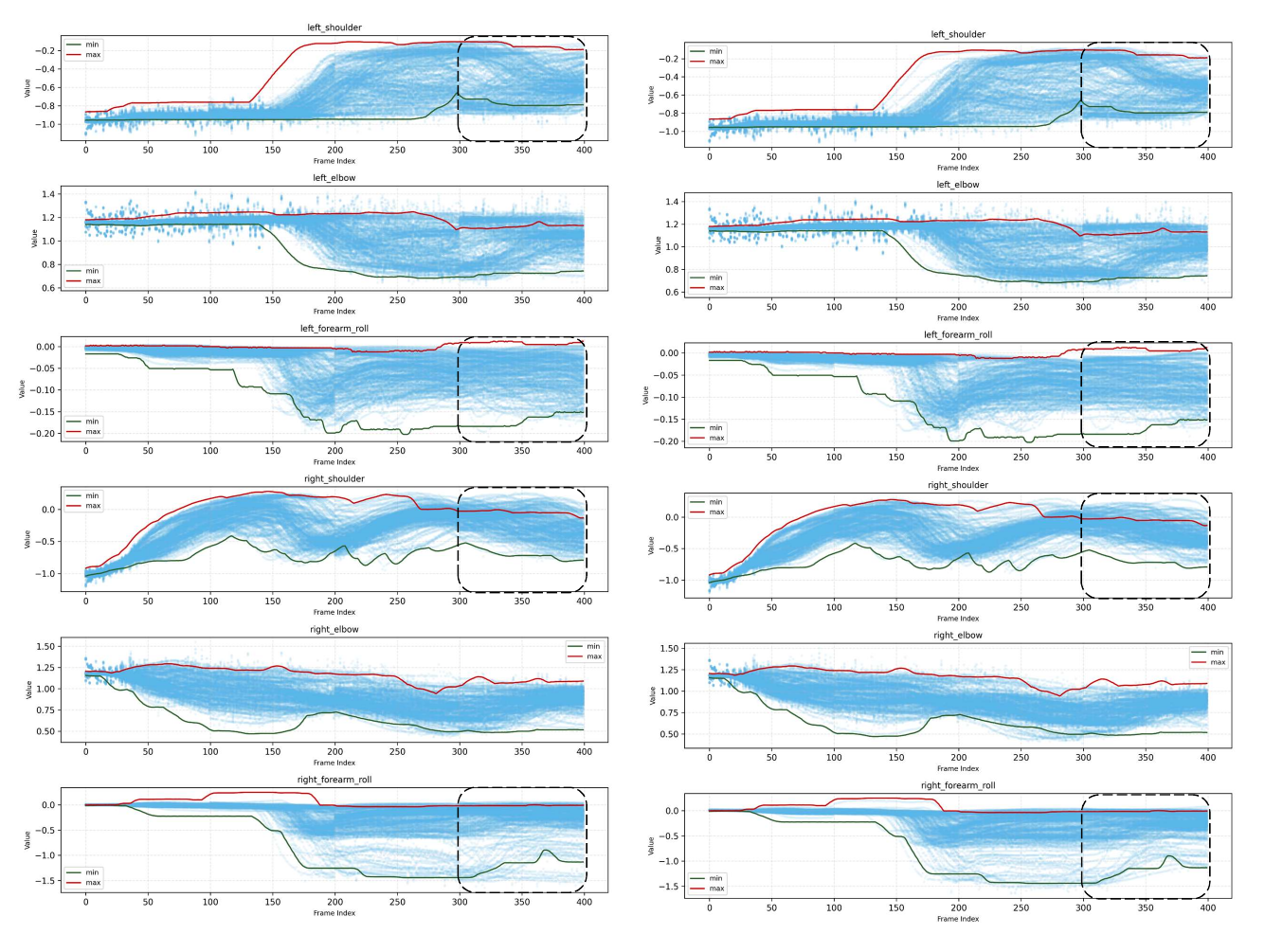}
\caption{High-priority Joint Group: (Left) ACT vs. (Right) ACT-ACORN}
\label{fig:priority_performance}
\end{figure}

\begin{figure}[h]
\centering
\includegraphics[width=\linewidth]{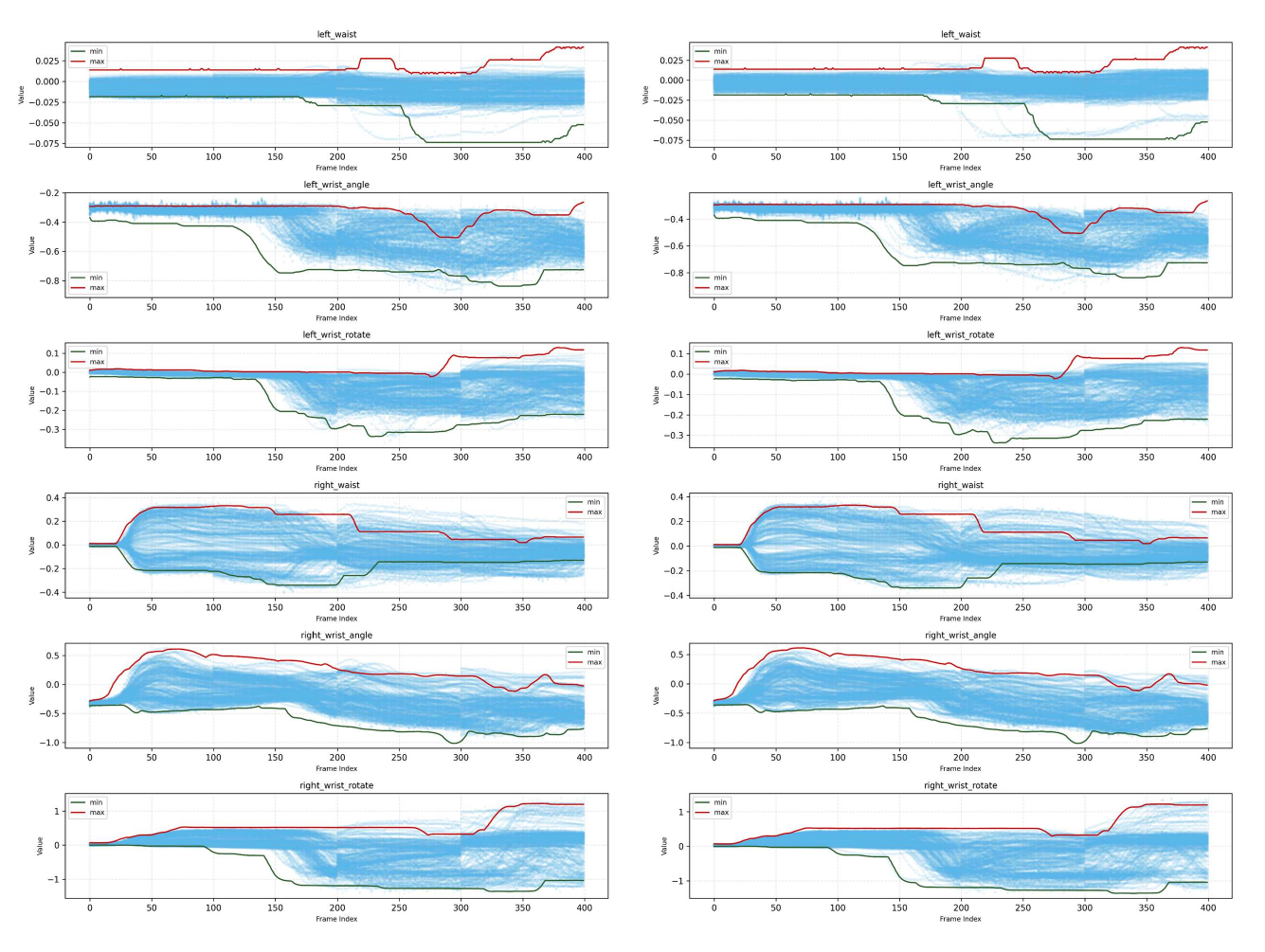}
\caption{Secondary Joint Group: (Left) ACT vs. (Right) ACT-ACORN}
\label{fig:secondary_performance}
\end{figure}

The results revealed three main findings in Fig~\ref{fig:priority_performance} and Fig~\ref{fig:secondary_performance}:

\begin{enumerate}

    \item For the high-priority joints, ACORN enhanced the convergence of the bilateral shoulder joints during the terminal movement phase, with a more concentrated region of trajectory density compared to baseline. 

    \item Under the control of ACORN, the rolling joints of both the left and right forearms showed excellent safety performance, with a substantial reduction in trajectories violating safety thresholds. 

    \item The secondary joint group showed slight but not significant differences in convergence and safety, which is consistent with our hypothesis that the improvement in fine manipulation stems primarily from optimized control of critical joints.

\end{enumerate}

The visualization results were able to confirm that ACORN was able to dynamically compensate for the joints most affected by fine manipulation.

\subsection{Ablation Studies}

\subsubsection{Curriculum Scheduling Coefficient $K$}

We first validate the curriculum learning mechanism through parameter ablation studies under NORMAL noise conditions. As shown in Table~\ref{tab:curriculum_steps}, progressively increasing $k$ demonstrates distinct performance patterns. The optimal configuration ($k=15$) improves success rate (SR) by \textbf{3.8\%} over $k=5$ baseline, while simultaneously AM-J and AM-E by \textbf{7.8\%} and \textbf{7.5\%} respectively. This empirically confirms our core hypothesis: gradual knowledge transfer from simple to complex tasks enables more stable skill acquisition.

\begin{table}[!htbp]
\centering
\caption{Ablation Study on Curriculum Scheduling Coefficient $K$}
\label{tab:curriculum_steps}
\small
\begin{tabular}{l *{5}{r}}
\toprule
\multirow{2}{*}{k} & \multicolumn{5}{c}{Performance Metrics} \\
\cmidrule(lr){2-6}
 & SR (\%) & ACR & ACR-F & AM-J & AM-E \\ 
\midrule
$5$  & 58.40  & 171.466 & 74.596 & 0.296 & 0.133 \\
$10$ & 61.00  & 173.930 & 71.149 & 0.296 & 0.137 \\
$15$ & \textbf{62.20} & \textbf{179.872} & \textbf{86.116} & \textbf{0.273} & \textbf{0.123} \\
\bottomrule
\end{tabular}
\end{table}

\begin{table}[!htbp]
\centering
\caption{Ablation Study on Contrastive Loss Coefficient \(\alpha\)}
\label{tab:margin_params}
\small
\begin{tabular}{l *{5}{r}}
\toprule
\multirow{2}{*}{$\alpha$} & \multicolumn{5}{c}{Performance Metrics} \\
\cmidrule(lr){2-6}
 & SR (\%) & ACR & ACR-F & AM-J & AM-E \\
\midrule
0.1   & \textbf{63.80} & \textbf{182.358} & 81.166  & 0.290 & \textbf{0.121} \\
0.01  & 62.20 & 179.872 & \textbf{86.116} & \textbf{0.273} & 0.123 \\
0.001 & 61.60 & 181.028 & 84.750 & 0.286 & 0.129 \\
\bottomrule
\end{tabular}
\end{table}

\subsubsection{Contrastive Loss Parameter \(\alpha\)}

Based on the identified optimal $k=15$ configuration, we further investigate margin parameter $\alpha$.  Table~\ref{tab:margin_params} reveals that $\alpha=0.01$ achieves the best trade-off between exploration and exploitation, particularly excelling in dynamic adaptation (ACR-F improvement of \textbf{6.1\%} over $\alpha=0.1$) while maintaining stable motion characteristics.

\section{Conclusions and Future Work}

In this work, we introduced ACORN, a novel strategy for enhancing safety and robustness in fine-grained robotic manipulation under environmental perturbations. ACORN addresses critical gaps in traditional imitation learning methods that prioritize performance over safety. Our key contributions include:

\begin{enumerate}
    \item \textbf{Safety-Centric Metric Ecosystem:} To address the critical gap in safety quantification within conventional embodied AI metrics, we introduce a multi-dimensional assessment system comprising four novel safety-oriented metrics. 
    \item \textbf{Plug-and-Play and Effective Deployment:} 
    ACORN achieves computationally efficient negative sampling through dual-domain perturbation mechanisms, eliminating dependency on additional data collection. It maintains architectural compatibility with existing policy networks, allowing it to be applied to a wide range of policies.
    \item \textbf{Adaptive Contrastive Optimization:} ACORN adaptively uses dynamic curriculum-aware comparative learning loss weights to ensure stable policy convergence. Achieving 23\% higher ACR-F and safer performance under moderate noise.
\end{enumerate}

Future research directions include extending its task generalization to heterogeneous robots, addressing sim-to-real transfer challenges, enabling dynamic environment adaptation, and integrating cross-modal sensing for contact-rich tasks. These advancements could establish ACORN as a foundational framework for real-world deployment.

\bibliographystyle{IEEEtranBST/IEEEtran}
\bibliography{main}

\end{document}